\documentclass[11pt, a4paper]{article}

\usepackage[utf8]{inputenc}
\usepackage[T1]{fontenc}

\usepackage[margin=2.5cm]{geometry}
\usepackage{graphicx}
\usepackage{amsmath}
\usepackage{amssymb}
\usepackage{booktabs} 
\usepackage{caption}
\usepackage{adjustbox}
\usepackage{multirow}
\usepackage{cellspace}
\usepackage{nicefrac}

\usepackage[dvipsnames,svgnames]{xcolor}
\usepackage{tikz}
\usepackage[edges]{forest} 
\usepackage{pgfplots}
\pgfplotsset{compat=1.18}
\usetikzlibrary{positioning, arrows.meta, chains, patterns}

\usepackage{hyperref}

\title{From Codicology to Code: A Comparative Study of Transformer and YOLO-based Detectors for Layout Analysis in Historical Documents}
\author{Sergio Torres Aguilar \\ {University of Luxembourg}}
\date{}

\begin{document}

\maketitle

\begin{abstract}
Robust Document Layout Analysis (DLA) is critical for the automated processing and understanding of historical documents with complex page organizations. This paper benchmarks five state-of-the-art object detection architectures on three annotated datasets representing a spectrum of codicological complexity: The e-NDP, a corpus of Parisian medieval registers (1326-1504); CATMuS, a diverse multiclass dataset derived from various medieval and modern sources (ca.12th-17th centuries) and HORAE, a corpus of decorated books of hours (ca.13th-16th centuries). We evaluate two Transformer-based models (Co-DETR, Grounding DINO) against three YOLO variants (AABB, OBB, and YOLO-World). Our findings reveal significant performance variations dependent on model architecture, data set characteristics, and bounding box representation. In the e-NDP dataset, Co-DETR achieves state-of-the-art results (0.752 mAP@.50:.95), closely followed by YOLOv11X-OBB (0.721). Conversely, on the more complex CATMuS and HORAE datasets, the CNN-based YOLOv11x-OBB significantly outperforms all other models (0.564 and 0.568, respectively). This study unequivocally demonstrates that using Oriented Bounding Boxes (OBB) is not a minor refinement but a fundamental requirement for accurately modeling the non-Cartesian nature of historical manuscripts. We conclude that a key trade-off exists between the global context awareness of Transformers, ideal for structured layouts, and the superior generalization of CNN-OBB models for visually diverse and complex documents.
\end{abstract}

\section{Introduction}

Historical documents are invaluable repositories of human history and knowledge. Unlocking their contents at scale requires sophisticated automated methods, among which Document Layout Analysis (DLA) plays a pivotal role \cite{binmakhashen2019survey}. DLA involves identifying and localizing distinct structural regions within a document page, such as text blocks, marginalia, page numbers, decorations, and signatures. This task is particularly challenging for manuscripts due to their inherent variability in script, layout conventions, page degradation, variable, and evolutive formats \cite{lombardi2020deep} as well as in early prints which normally follow manuscript layouts and graphic styles. Accurate DLA is a prerequisite for subsequent tasks such as optical character recognition (OCR), handwritten text recognition (HTR), and content-based information retrieval in order to well establish different page units and transcribe and interpret content following their internal logic.

Although traditional DLA methods relied on rule-based systems or classical image processing, deep learning has become the dominant paradigm. Object detection techniques, initially developed for natural or medical images, have been successfully adapted for DLA. Early deep learning approaches often employed two-stage detectors like Faster R-CNN \cite{ren2015faster}. More recently, efficient single-stage detectors, particularly the You Only Look Once (YOLO) family, whose 11 version is used in this paper, \cite{redmon2016yolo}, have gained prominence because of their speed-accuracy trade-off.

Transformer architectures are increasingly applied to vision tasks. End-to-end detectors like DINO \cite{zhang2022dino} and Co-DETR \cite{zong2023detrs} eliminate the need for hand-crafted components such as Non-Maximum Suppression (NMS) and offer strong performance, leading the board on mAP for vast datasets like COCO \cite{lin2014microsoft} and Objects365 \cite{shao2019objects365}. Moreover, Vision-Language Models (VLMs) trained on large-scale image-text pairs have opened new possibilities. Grounding DINO \cite{liu2024grounding} and hybrid methods like YOLO-World \cite{cheng2024yolo} integrate language understanding for open-vocabulary detection, enabling object specification via text prompts.

Given this diverse landscape of high-performance models, selecting the optimal architecture for DLA for specific historical document types remains an open question. This paper aims to bridge this gap through a rigorous comparison of five representative modern object detection models fine-tuned on three distinct historical manuscript datasets. We evaluate:
\begin{itemize}
    \item \textbf{Grounding DINO (GD):} An VLM with strong grounding capabilities (open vocabulary).
    \item \textbf{Co-DETR:} Advanced end-to-end Transformer detector.
    \item \textbf{YOLOv11x (AABB):} Efficient CNN detectors with axis-aligned outputs.
    \item \textbf{YOLOv11x (OBB):} Efficient CNN detectors adapted for oriented outputs, which are hypothesized to be beneficial for skewed document elements.
    \item \textbf{YOLO-Worldx8x (YOLOW):} Hybrid architectures aimed at real-time (open vocabulary detection).
\end{itemize}

We analyze their performance on the relatively homogeneous e-NDP dataset and the more diverse CATMuS and HORAE datasets. Our contributions include: (1) benchmarking SOTA models on specific historical DLA tasks, (2) quantifying the impact of using Oriented Bounding Boxes (OBB) for YOLO models in this domain, (3) designing and validating a data harmonization methodology based on a hierarchical, codicologically-informed ontology, and using it to train a robust foundational model (`YOLO-gen`), and (4) discussing the strengths and weaknesses of different architectural paradigms based on empirical results across datasets of varying complexity.

\section{Related Work}

Document Layout Analysis has been a long-standing challenge in document image processing. Early methods relied on rule-based systems \cite{nagy2005state} or projection profiles \cite{o1995document}, often effective for structured documents but brittle for historical manuscripts. Machine learning approaches using hand-crafted features and decision trees followed \cite{im2005change}.

Deep learning revolutionized DLA. Convolutional neural networks (CNNs) have formed the backbone of many successful models. Two-stage detectors such as Faster R-CNN \cite{ren2015faster}, sometimes adapted specifically for documents \cite{mijwil2022distinction}, provide robust baseline performance. Single-stage detectors, particularly YOLO variants \cite{redmon2016yolo}, offer faster alternatives, crucial for large-scale processing, although sometimes at the cost of accuracy on complex layouts \cite{diwan2023yolo}. 

The advent of Transformers led to end-to-end object detectors such as DETR \cite{carion2020detr}, which eliminated the need for anchors and NMS. Subsequent improvements like Deformable DETR \cite{zhu2021deformable}, DINO \cite{zhang2022dino}, and Co-DETR \cite{zong2023detrs} further enhanced performance and efficiency. Co-DETR, specifically, introduces cooperative training between the detection head and auxiliary heads to improve feature learning. These models have shown promise for DLA tasks that require an understanding of the global context \cite{li2022dit}.

Simultaneously, Vision-Language Models (VLMs) pre-trained on large image-text corpora emerged. Models like CLIP \cite{radford2021clip} provided powerful image representations aligned with text. Subsequent works such as GLIP \cite{li2022glip} and Grounding DINO \cite{liu2024grounding} extended this to open-set object detection, allowing detection based on free-text prompts. Their ability to leverage semantic understanding makes them attractive for DLA where elements are often defined semantically. YOLO-World \cite{cheng2024yolo} attempts to merge the efficiency of YOLO with these open-vocabulary capabilities. Florence-2 \cite{xiao2024florence2} represents another powerful VLM designed for multi-task vision capabilities via sequence-to-sequence learning using specific task prompts and quantized location tokens.

A specific challenge in DLA, particularly for historical documents and text detection, is handling non-axis-aligned elements. This has led to the development of orientation bounding box (OBB) detection methods \cite{wen2023comprehensive}. Frameworks like MMRotate \cite{zhou2022mmrotate} provide tools and models specifically for this purpose, adapting architectures like Faster R-CNN and RetinaNet. YOLO models have also seen OBB adaptations \cite{sun2021bifa}, as tested in this work. Recent works like DocLayout-YOLO \cite{zhao2024doclayoutyolo} also highlight optimizations for DLA within the YOLO paradigm.

Our work contributes by directly comparing leading fine-tunable models from these different paradigms (CNN-AABB, CNN-OBB, Transformer-DETR, Transformer-VLM, Hybrid-VL-YOLO) on the specific domain of medieval and modern manuscript DLA across three datasets, providing valuable empirical benchmarks for practitioners.


\section{Datasets and Experimental Setup}

\subsection{Datasets}

We utilized three distinct datasets for fine-tuning and evaluation:

\subsubsection{e-NDP Dataset}

Derived from the Electronic Notre-Dame de Paris project \cite{claustre2023endp}, this dataset consists of images from 26 medieval registers of the Parisian Chapter's deliberations (1326-1504). The layout exhibits a relatively structured evolution over time. There are five different annotated key semantic classes without overlapping. These classes were identified as relevant in previous work \cite{claustre2023endp}:
\begin{itemize} \itemsep0em 

    \item \textbf{Page Number}: Roman or Arabic numerals, typically in corners. 
    \item \textbf{Columnar Name List}: Vertical names lists of meeting assistants.
    \item \textbf{Primary Text Region}: Blocks of main text (meeting conclusions).
    \item \textbf{Marginal Index Notes}: Notes/entries in margins related to the main text. 
    \item \textbf{Date Line}: Lines containing the date, often preceding the main text.
\end{itemize}
Original annotations were polygon-based PAGE XML. These were converted into COCO AABB and YOLO OBB formats. The data set was divided 90\% / 10\% for the trainval / test, resulting in a test set of 37 images with 275 instances. See Table \ref{tab:class_dist_compact}

\begin{figure}[h]
\centering

\scalebox{0.17}{
\includegraphics{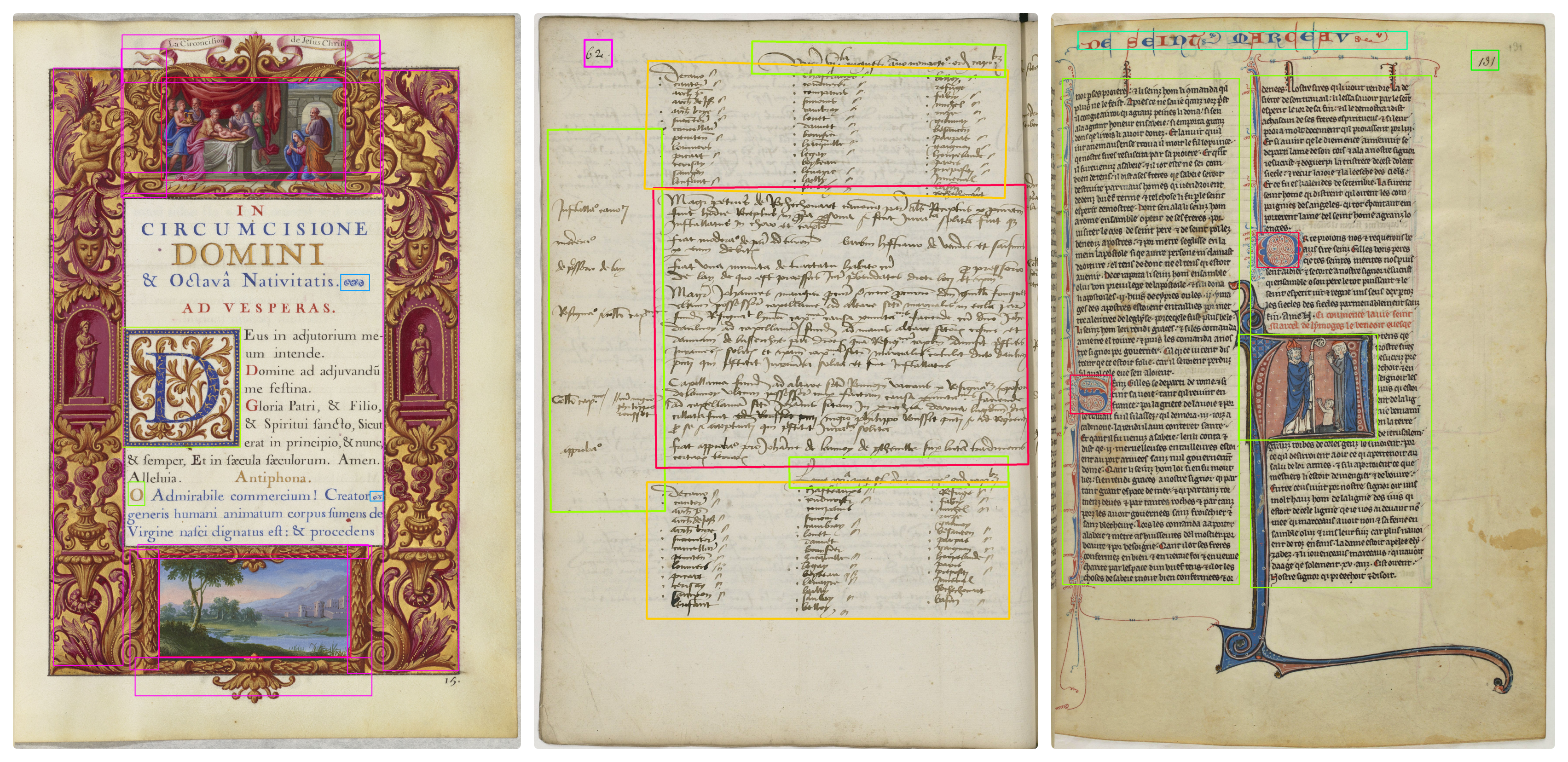}}
\centering
\caption[foo bar]{\small{Three annotated pages from the HORAE (pray book), e-NDP (registers) and CATmUS (litterature) datasets respectively.}

   }
\label{fig:mufi_stats}
\end{figure}

\subsubsection{CATMuS Dataset}

Sourced from the Hugging Face dataset `CATMuS/medieval-segmentation` \cite{constum2024catmus}, which contains diverse annotations from various medieval manuscripts (administrative, literary, early printed). To create a data set focused on region-level detection, we applied the following pre-processing:

\begin{enumerate}
    \item Line-level annotations were excluded. 
    \item Only categories that appeared 1 or more times on the test split were retained. 
    \item This filtering resulted in the following 9 valid region categories used in this study: 
        \begin{itemize} \itemsep0em 
            \item \textbf{DropCapitalZone:} Zone containing a large ornamental initial letter.
            \item \textbf{GraphicZone:} Zone with non-textual elements (images, drawings, decorations).
            \item \textbf{MainZone:} The main textual body zone of a page.
            \item \textbf{MarginTextZone:} Text zone in the margin (annotations, commentary, etc.).
            \item \textbf{NumberingZone:} Zone containing page, folio, or other numbering.
            \item \textbf{QuireMarksZone:} Zone containing quire or gathering marks.
            \item \textbf{RunningTitleZone:} Zone containing a running title (typically top of the page).
            \item \textbf{StampZone:} Zone containing a stamp (library, ownership, etc.).
            \item \textbf{RunningTitleZone: } Running title at the top of the page.
        \end{itemize}
    \item The raw category names (e.g., `DropCapitalZone`) were mapped to descriptive phrases (e.g., "ornate drop capital letter zone") for use in VLM training configurations and output metadata, controlled by a script flag. 

\end{enumerate}

\subsubsection{HORAE Dataset}

This dataset was created as part of the HORAE project \cite{boillet2019horae}. It is composed of pages from Books of Hours, best-selling personal prayer books in the late Middle Ages, known for their lavish illustrations. They represent a significant DLA challenge due to complex layouts featuring intricate borders, miniatures, and a wide variety of initials. We focus on 8 distinct layout classes that include decorative and structural elements.

\begin{itemize} \itemsep0em 
    \item \textbf{Decorated Border}: Margins filled with ornamental vine work or geometric patterns.
    \item \textbf{Illustrated Border}: Margins containing identifiable figures, scenes, or objects.
    \item \textbf{Miniature}: A full-page or large-scale illustration depicting a narrative scene.
    \item \textbf{Decorated Initial}: An enlarged initial letter with non-figurative ornamentation.
    \item \textbf{Historiated Initial}: An enlarged initial letter containing an identifiable scene or figure.
    \item \textbf{Line Filler}: A decorative element filling the end of a line of text.
    \item \textbf{Simple Initial}: A simple colored or larger initial letter without significant decoration.
    \item \textbf{Border Text}: Text written within the marginal space, often as captions or glosses.
\end{itemize}

\subsection{Experimental Setups}

For this comparative study the data sets were processed into standard COCO AABB format for models requiring axis-aligned bounding boxes and the YOLO OBB format (using minimum area rectangles derived from source polygons) for the oriented bounding box detector. A 90\%/10\% split was used to create trainval/test sets. The key characteristics are summarized in Table \ref{tab:dataset_summary}.

\begin{table}[htbp!]
\centering
\caption{Summary of the Datasets Used for Fine-tuning and Evaluation.}
\label{tab:dataset_summary}
\scalebox{0.84}{ 
\begin{tabular}{lccc}
\toprule
\textbf{Feature} & \textbf{e-NDP} & \textbf{CATMuS} & \textbf{HORAE} \\
\midrule
Original Source & `e-NDP` (Zenodo) & `CATMuS` (HF) & `HORAE` (Github) \\
Document Types & Administrative Registers & Diverse (Admin., Lit., Prints) & Books of Hours (Devotional) \\
Layout Type & Relatively Homogeneous & Moderately Diverse & Visually Rich \& Complex \\
Annotation Source & PAGE XML Polygons & COCO format & PAGE XML Polygons \\
Preprocessing & 5 types to Classes & Classes from test set & Classes with instances \\
\textbf{Classes Used} & \textbf{5} & \textbf{9} & \textbf{8} \\
Trainval Images & 327 & 1525 & 418 \\
Test Images & 37 & 158 & 45 \\
Test Instances & 275 & 733 & 558 \\
Output Formats & \small{COCO AABB, YOLO OBB} & \small{COCO AABB, YOLO OBB} & \small{COCO AABB, YOLO OBB} \\
\bottomrule
\end{tabular}}
\end{table}

The specific layout categories targeted in each dataset differ due to their origins and the absence of a community-defined ontology. However, despite different naming conventions, several classes represent similar layout concepts (e.g., main text block, page number, capitals and marginalia) as they are in general inspired on the codicological vocabulary from Muzerelle \cite{muzerelle-cod}.

\begin{table}[htbp!]
\centering
\caption{Class instance counts in the test sets for the e-NDP, CATMuS, and HORAE datasets.}
\label{tab:class_dist_compact}
\scalebox{0.95}{ 
\begin{tabular}{lr@{\qquad}lr@{\qquad}lr}
\toprule
\multicolumn{2}{c}{\textbf{e-NDP (Total: 275)}} & \multicolumn{2}{c}{\textbf{CATMuS (Total: 733)}} & \multicolumn{2}{c}{\textbf{HORAE (Total: 558)}} \\
\cmidrule(r){1-2} \cmidrule(lr){3-4} \cmidrule(l){5-6}
Class & Count & Class & Count & Class & Count \\
\midrule
Primary Text Region  & 91 & MainZone         & 275 & Decorated Initial & 282 \\
Page Number          & 57 & MarginTextZone   & 199 & Line Filler       & 112 \\
Date Line            & 54 & DropCapitalZone  & 124 & Decorated Border  & 77  \\
Columnar Name List   & 48 & NumberingZone    & 94  & Simple Initial    & 57  \\
Marginal Index Notes & 25 & RunningTitleZone & 18  & Miniature         & 15  \\
                     &    & GraphicZone      & 10  & Illustrated Border& 10  \\
                     &    & QuireMarksZone   & 8   & Border Text       & 4   \\
                     &    & StampZone        & 4   & Historiated Initial& 1  \\
                     &    & TitlePageZone    & 1   &                   &     \\
\bottomrule
\end{tabular}}
\end{table}

\subsection{Models Compared}

We evaluated five distinct model families, selecting representative variants and using publicly available pre-trained weights where possible. A summary is provided in Table \ref{tab:model_summary}.

\begin{table}[htbp]
\centering

\caption{Summary of Compared Object Detection Models}
\label{tab:model_summary}
\scalebox{0.94}{
\begin{tabular}{ Sc Sc Sc Sc Sc Sc}
\toprule
Full Name         & Architecture & Key Components                  & Pre-training  & Input   & Par (M)    \\
\midrule
Grounding DINO    & VLM-Transformer   & Swin-T + BERT & Obj365/GoldG & \small{AABB}     & 172   \\
YOLOv11x        & CNN (1-Stage)     & \small{CNN + C3k2/C2PSA}   & COCO      & \small{AABB}     & 57   \\
YOLOv11x-obb& CNN (1-Stage)     & \small{YOLO + OBB Head}      & COCO      & \small{OBB}      & 57   \\
YOLOv8x-world          & VL-YOLO    & \small{YOLO + CLIP}     & Obj365/GoldG    & \small{AABB}     & 110  \\
Co-DETR    & DETR-Variant      & CNN + Swin-L  & COCO/Obj365   & \small{AABB}     & 218   \\
\bottomrule
\end{tabular}}
\end{table}


\subsection{Cross-Corpus Training: The YOLO-gen Foundational Model}

To move beyond single-corpus benchmarks, our final experiment addresses a core challenge in applying machine learning to historical documents: the lack of standardized annotation ontologies. Datasets like e-NDP, CATMuS, and HORAE, while invaluable, use particularized class schemes that are not directly interoperable. This experiment, therefore, tests a methodology for harmonizing these disparate datasets by creating a unified, hierarchical model.

The cornerstone of this approach is a \emph{codicologically-informed reclassification} of the three source ontologies. While the specific tags in each corpus are distinct, they often refer to common, fundamental codicological elements. We created a unified ontology to capture this shared vocabulary, establishing high-level parent categories from scholarly concepts (\texttt{Text}, \texttt{Paratext}, \texttt{Initials}, etc.) and mapping original tags as subclasses  within this structure. This provides a coherent framework for low-conflict training.

The resulting ontology, visualized in Figure \ref{fig:ontology_tree_final}, was built upon three key principles:

\begin{enumerate}
    \item \textbf{Hierarchical Structure:} Instead of a flat list of classes, we defined a multi-level hierarchy with parent, intermediate, and subclass categories to represent the relationships between layout elements.
    \item \textbf{Multi-Label Training:} Each annotated object was assigned multiple labels corresponding to its full path in the hierarchy. For instance, a simple initial from the HORAE dataset was tagged as \texttt{Initial} (Level 1 parent), \texttt{Initial\_Manuscript} (Level 2 intermediate parent), and \texttt{Initial\_Ms\_Simple} (Level 3 subclass). This provides a richer training signal.
    \item \textbf{Semantic and Visual Coherence:} The ontology was iteratively refined to balance functional groupings (e.g., creating a \texttt{Paratext} class for non-primary text) with visual distinctions (e.g., separating \texttt{Initial\_Manuscript} from \texttt{Initial\_Printed}) to aid the model's learning process.
\end{enumerate}

The best model trained on this dataset, which we term \textbf{YOLO-gen}, was hypothesized to develop a robust comprehension of the abstract parent categories, even if its performance on specific subclasses might not surpass that of the specialized models. This would validate YOLO-gen as a strong baseline for future research projects requiring codicological segmentation.

The performance of this unified model is compared against the specialist models in the following sections. Table \ref{tab:yolo_gen_parents} shows the performance of our \textbf{YOLO-gen} model on the abstract parent classes, a task that the specialist models cannot perform.

\definecolor{color_endp}{HTML}{D9534F}   
\definecolor{color_catmus}{HTML}{5CB85C} 
\definecolor{color_horae}{HTML}{F0AD4E}  

\newcommand{\sourceE}{\textcolor{color_endp}{\scriptsize\bfseries\sffamily E}}
\newcommand{\sourceC}{\textcolor{color_catmus}{\scriptsize\bfseries\sffamily C}}
\newcommand{\sourceH}{\textcolor{color_horae}{\scriptsize\bfseries\sffamily H}}

\begin{figure}[ht]
\centering
\caption{Hierarchical structure for YOLO-gen merging the three corpus annotations. The colored initials indicate the original corpus for each class: \sourceE{} (e-NDP), \sourceC{} (CATMuS), \sourceH{} (HORAE).}
\label{fig:ontology_tree_final}
\begin{forest}
  for tree={
    grow=east,
    draw,
    thick,
    rounded corners,
    font=\sffamily,
    align=center, 
    l sep=1.1cm,
    s sep=1.5mm,
    inner ysep=3pt,
    inner xsep=4pt,
    forked edges,
    edge={-Latex, thick, color=black!60},
    where level=0{
      font=\small\bfseries,
      fill=black!14,
    }{},
    where level=1{
      font=\small\sffamily\bfseries,
      fill=blue!15,
      text width=2cm, 
    }{},
    where level=2{
      font=\footnotesize\sffamily,
      fill=blue!7,
      text width=2.8cm,
    }{},
    where level=3{
      font=\scriptsize\sffamily,
      fill=gray!12,
      text width=3cm,
    }{},
  }
  [Unified\\Ontology
    [Text
      [Text\_Main, label=below:{\sourceE, \sourceC}]
      [Paratext
        [Paratext\_Marginal, label=right:{\sourceE, \sourceC, \sourceH}]
        [Paratext\_Header, label=right:{\sourceC}]
        [Paratext\_List, label=right:{\sourceE}]
        [Paratext\_DateLine, label=right:{\sourceE}]
      ]
    ]
    [Decoration
        [Deco\_Border, label={[xshift=1mm]right:{\sourceH}}]
        [Deco\_Miniature, label=right:{\sourceH}]
        [Deco\_Graphic, label=right:{\sourceC}]
        [Deco\_LineFiller, label={[yshift=-1.5mm]right:{\sourceH}}]
    ]
    [Initial
      [Initial\_Manuscript
          [Initial\_Ms\_Simple, label=right:{\sourceH}]
          [Initial\_Ms\_Decorated, label=right:{\sourceH}]
          [Initial\_Ms\_Historiated, label=right:{\sourceH}]
      ]
      [Initial\_Printed
          [Initial\_P\_DropCapital, label=right:{\sourceC}]
      ]
    ]
    [Numbering
        [Numbering\_Page, label=right:{\sourceE, \sourceC}]
    ]
    [Marks
        [Marks\_Quire, label=right:{\sourceC}]
        [Marks\_Stamp, label=right:{\sourceC}]
        [Marks\_Seal, label=right:{\sourceC}]
    ]
    [Damage
        [Damage\_Generic, label=right:{\sourceC}]
        [Damage\_Scan, label=right:{\sourceC}]
    ]
  ]
\end{forest}
\end{figure}
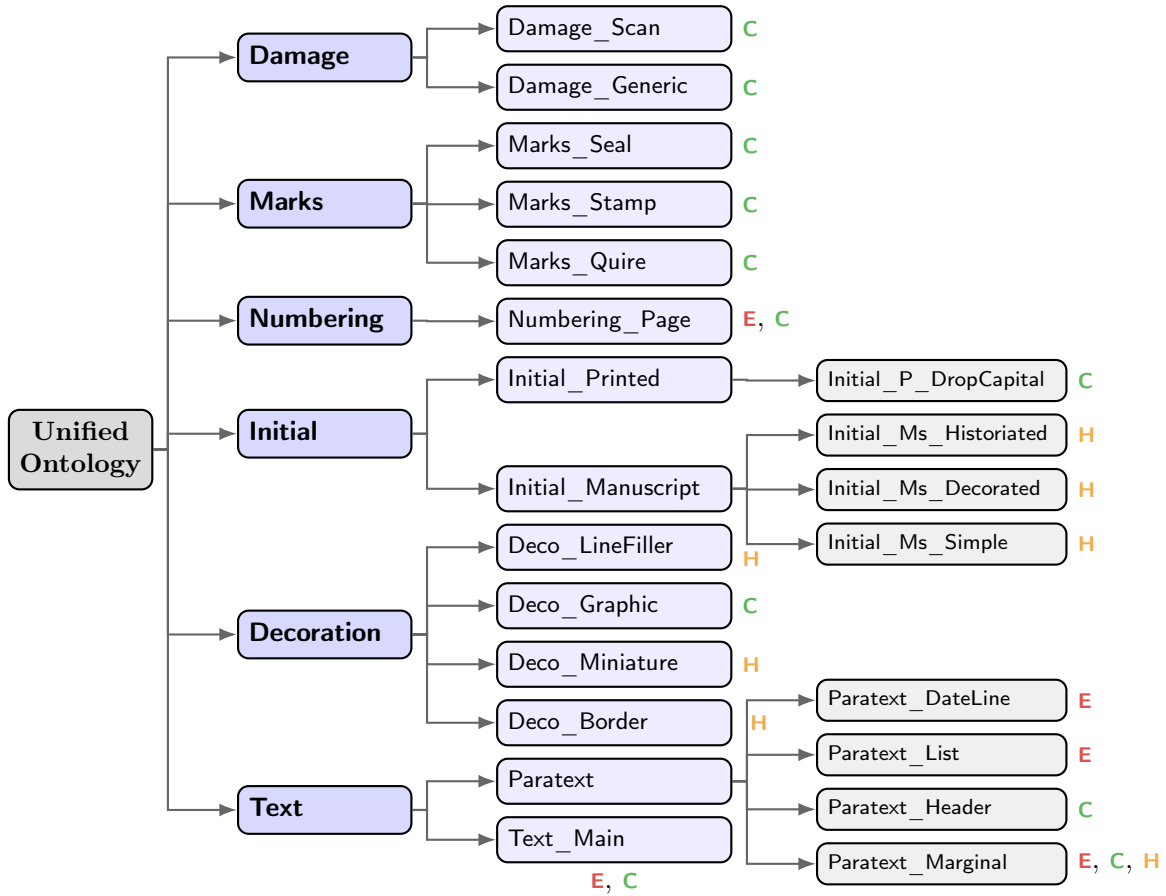

\begin{table}[htbp!]
\centering
\caption{Performance of the YOLO-gen 11x-OBB model on the abstract parent and intermediate classes, evaluated on the combined test set.}
\label{tab:yolo_gen_parents}
\scalebox{0.98}{
\begin{tabular}{lcccc}
\toprule
\textbf{Parent/Intermediate Class} & \textbf{mAP@.50:.95} & \textbf{mAP@.50} & \textbf{Precision} & \textbf{Recall} \\
\midrule
Text & 0.675 & 0.861 & 0.749 & 0.861 \\
Decoration & 0.629 & 0.839 & 0.712 & 0.902 \\
Initial (Universal) & 0.662 & 0.880 & 0.748 & 0.878 \\
Marks & 0.665 & 0.821 & 0.643 & 0.900 \\
Numbering & 0.422 & 0.776 & 0.611 & 0.820 \\
Paratext (Intermediate) & 0.461 & 0.674 & 0.643 & 0.658 \\
Initial\_Manuscript (Inter.) & 0.416 & 0.519 & 0.801 & 0.225 \\
Initial\_Printed (Inter.) & 0.477 & 0.720 & 0.755 & 0.597 \\
\midrule
Average  & 0.551 & 0.761 & 0.701 & 0.730 \\
\bottomrule
\end{tabular}}
\end{table}

\subsection{Fine-tuning and Evaluation}
Each model was independently fine-tuned on the 'trainval' split of each data set for each corpus typology (AABB or OBB) and evaluated on the corresponding 'test' split. We report the results from the checkpoint that yielded the best mAP@.5:.95 on the test set. MMDetection and Ultralytics frameworks were used, enabling standard hyperarameters. The evaluation strictly followed the standard COCO bounding box evaluation protocol using mAP@0.50:0.95 (mAP) and mAP@0.50 (mAP50), calculated overall and per class. All models were trained for a minimum of 100 epochs with a consistent image size of 640x640 using a Cosine scheduler.

\section{Results}

\subsection{Results on e-NDP Dataset (5 Classes)}

Table \ref{tab:overall_endp_final} summarizes the overall performance metrics for all five models in the e-NDP test set. Tables \ref{tab:per_class_map_endp_final} and \ref{tab:per_class_map50_endp_final} provide the breakdown per category.

\begin{table}[ht]
\centering
\caption{Overall Performance Comparison on e-NDP Dataset (5 Classes)}
\label{tab:overall_endp_final}
\scalebox{0.97}{ 
\begin{tabular}{lccccc}
\toprule
Model /                  & mAP                     & mAP                    & Recall                 & Precision             & Speed \\
 Epochs                      & @.50:.95 ($\uparrow$)    & @.50 ($\uparrow$)      & (AR/R) ($\uparrow$)    & (P) ($\uparrow$)      & (\nicefrac{\text{im}}{\text{ms}})$^\dagger$ ($\downarrow$) \\
\midrule
Grounding DINO (ep22)  & 0.718          & 0.954          & 0.812 (AR)     & N/A            & $\approx$102.14         \\
YOLOv11x-AABB (ep67)   & 0.633          & 0.893          & 0.836 (R)      & 0.861          & $\approx$7.4            \\
YOLOv11x-OBB (ep88)    & 0.721          & 0.941          & \textbf{0.892} (R) & \textbf{0.902} & $\approx$11.1           \\
YOLO-World x (ep70)    & 0.576          & 0.880          & 0.805 (R)      & 0.830          & $\approx$12.6           \\
Co-DETR (ep14)         & \textbf{0.752} & \textbf{0.979} & 0.810 (AR)     & N/A            & $\approx$140.0          \\
\midrule
YOLO-gen 11x-OBB (ep88)   & 0.673 & 0.892 & 0.836 (Avg) & 0.863 (Avg) & $\approx$14.4  \\
\bottomrule
\end{tabular}
} 
\caption*{\footnotesize $^\dagger$Inference times estimated on NVIDIA A6000 (Batch Size=1). AR/R indicates Average Recall (MMDet) or Recall (Ultralytics).}
\end{table}

\begin{table}[htbp!]
\centering
\caption{Per-Class mAP@0.50:0.95 Comparison on e-NDP Dataset (5 Classes)}
\label{tab:per_class_map_endp_final}

\scalebox{0.99}{ 
\begin{tabular}{lcccccc}
\toprule

Models /                  & \small{Grounding}                     & \small{YOLO}                    & \small{YOLO}                 & \small{YOLO}             & \small{Co-DINO} & \small{YOLO-gen} \\
 Category                      & \small{DINO}    & \small{11x-AABB}      & \small{11x-OBB}    & Worldx (v2)  & \small{(co-DETR)} & \small{11x-OBB} \\
\midrule

Page Number         & 0.469 & 0.390 & \textbf{0.605} & 0.377 & 0.537 & 0.530 \\
Col. Name List      & 0.827 & 0.774 & 0.828 & 0.751 & \textbf{0.850} & 0.828 \\
Primary Text        & 0.850 & 0.724 & 0.749 & 0.695 & \textbf{0.862} & 0.719 \\
Marginal Notes      & 0.722 & 0.643 & 0.732 & 0.544 & \textbf{0.762} & 0.647 \\
Date Line           & 0.724 & 0.632 & 0.689 & 0.511 & \textbf{0.749} & 0.642 \\

\bottomrule
\end{tabular}}

\end{table}

\begin{table}[htbp!]
\centering
\caption{Per-Class mAP@0.50 Comparison on e-NDP Dataset (5 Classes)}
\label{tab:per_class_map50_endp_final}
\scalebox{0.99}{ 
\begin{tabular}{lcccccc}
\toprule

Models /                  & \small{Grounding}                     & \small{YOLO}                    & \small{YOLO}                 & \small{YOLO}             & \small{Co-DINO} & \small{YOLO-gen} \\
 Category                      & \small{DINO}    & \small{11x-AABB}      & \small{11x-OBB}    & Worldx (v2)  & \small{(co-DETR)} & \small{11x-OBB} \\
\midrule
Page Number         & 0.946 & 0.849 & 0.925 & 0.815 & \textbf{0.959} & 0.855 \\
Col. Name List      & \textbf{0.998} & 0.941 & 0.967 & 0.946 & \textbf{0.998} & 0.937 \\
Primary Text        & 0.957 & 0.890 & 0.935 & 0.884 & \textbf{0.966} & 0.908 \\
Marginal Notes      & 0.885 & 0.866 & 0.943 & 0.820 & \textbf{0.988} & 0.849 \\
Date Line           & 0.983 & 0.920 & 0.936 & 0.934 & \textbf{0.985} & 0.912 \\

\bottomrule
\end{tabular}}
\end{table}

In the relatively homogeneous e-NDP dataset, the Transformer-based data set \textbf{Co-DETR stands out}, achieving the highest overall mAP50:95 (0.752) and mAP50 (0.979), as well as the best AP75 (0.838). It also leads in performance for nearly all individual categories. \textbf{YOLOv11l-OBB} and \textbf{Grounding DINO} closely follow with high mAP scores (0.721 and 0.718 respectively), significantly outperforming YOLOv11l-AABB (0.633) and YOLO-World (0.576). This highlights the effectiveness of both advanced Transformer architectures and OBB-enhanced CNNs for this specific structured layout task. YOLOv11l-OBB notably achieves the highest Recall.

\subsection{Results on CATMuS Dataset (Filtered - 9 Classes)}

Performance in the more diverse, filtered CATMuS dataset is summarized in Table \ref{tab:overall_catmus_final}. The per-class results for the common categories evaluated are in Tables \ref{tab:per_class_map_catmus_final} and \ref{tab:per_class_map50_catmus_final}.

\begin{table}[ht]
\centering
\caption{Overall Performance Comparison on CATMuS Dataset (Filtered - 9 Classes)}
\label{tab:overall_catmus_final}
\scalebox{1}{
\begin{tabular}{lccccc}
\toprule
Model /                  & mAP                     & mAP                    & Recall                 & Precision             & Speed \\
 Epochs                      & @.50:.95 ($\uparrow$)    & @.50 ($\uparrow$)      & (AR/R) ($\uparrow$)    & (P) ($\uparrow$)      & (\nicefrac{\text{im}}{\text{ms}}) ($\downarrow$) \\
\midrule
Grounding DINO (ep30)   & 0.452          & 0.706          & 0.622 (AR)     & N/A   & $\approx$111.2         \\
YOLOv11x-AABB (ep76)  & 0.425          & 0.631          & 0.544 (R)       & 0.607 & $\approx$7.5         \\
YOLOv11x-OBB (ep91)   & \textbf{0.564} & \textbf{0.740} & 0.663 (R) & \textbf{0.718} & $\approx$11.2 \\
YOLO-World x (ep67)   & 0.425          & 0.682          & 0.631 (R)       & 0.619          & $\approx$12.6\\
Co-DETR (ep16)          & 0.492          & 0.737          & 0.643 (AR)  & N/A    & $\approx$147.1 \\
\midrule
YOLO-gen 11x-OBB (ep88)   & 0.560 & 0.735 & \textbf{0.698} (R) & 0.640 & $\approx$13.8   \\
\bottomrule
\end{tabular}}
\end{table}

\begin{table}[htbp]
\centering
\caption{Per-Class mAP@0.50:0.95 Comparison on CATMuS Dataset (Common 9 Classes)}
\label{tab:per_class_map_catmus_final}
\scalebox{0.97}{ 
\begin{tabular}{lcccccc}
\toprule

Models /                  & \small{Grounding}                     & \small{YOLO}                    & \small{YOLO}                 & \small{YOLO}             & \small{Co-DINO} & \small{YOLO-gen} \\
 Category                      & \small{DINO}    & \small{11x-AABB}      & \small{11x-OBB}    & Worldx (v2)  & \small{(co-DETR)} & \small{11x-OBB} \\
\midrule

DropCapitalZone     & 0.339 & 0.291 & \textbf{0.494} & 0.340 & 0.348 & 0.485 \\
GraphicZone         & \textbf{0.750} & 0.547 & 0.709 & 0.739 & 0.695 & 0.622 \\
MainZone            & 0.861 & 0.846 & 0.861 & 0.859 & \textbf{0.868} & 0.840 \\
MarginTextZone      & 0.304 & 0.163 & 0.313 & 0.231 & 0.260 & \textbf{0.338} \\
NumberingZone       & \textbf{0.145} & 0.102 & \textbf{0.349} & 0.121 & 0.158 & 0.215 \\
QuireMarksZone      & 0.483 & 0.394 & 0.709 & 0.348 & 0.501 & \textbf{0.752} \\
RunningTitleZone    & 0.129 & 0.089 & \textbf{0.175} & 0.067 & 0.120 & 0.115 \\
StampZone           & 0.606 & 0.499 & 0.573 & 0.424 & 0.575 & \textbf{0.864} \\
TitlePageZone       & 0.400 & 0.895 & \textbf{0.895} & 0.697 & 0.900 & 0.810 \\

\bottomrule
\end{tabular}
}

\end{table}

\begin{table}[htbp!]
\centering
\caption{Per-Class mAP@0.50 Comparison on CATMuS Dataset (Common 9 Classes)}
\scalebox{0.97}{ 
\label{tab:per_class_map50_catmus_final}
\begin{tabular}{lcccccc}
\toprule

Models /                  & \small{Grounding}                     & \small{YOLO}                    & \small{YOLO}                 & \small{YOLO}             & \small{Co-DINO} & \small{YOLO-gen} \\
 Category                      & \small{DINO}    & \small{11x-AABB}      & \small{11x-OBB}    & Worldx (v2)  & \small{(co-DETR)} & \small{11x-OBB} \\
\midrule

DropCapitalZone     & 0.617 & 0.556 & 0.693 & 0.682 & 0.645 & \textbf{0.730} \\
GraphicZone         & \textbf{0.815} & 0.694 & 0.803 & 0.803 & 0.769 & 0.677 \\
MainZone            & \textbf{0.953} & 0.951 & 0.949 & 0.951 & \textbf{0.953} & 0.931 \\
MarginTextZone      & \textbf{0.637} & 0.351 & 0.544 & 0.500 & 0.573 & 0.576 \\
NumberingZone       & 0.500 & 0.339 & \textbf{0.733} & 0.394 & 0.551 & 0.503 \\
QuireMarksZone      & 0.945 & 0.731 & 0.955 & 0.895 & 0.950 & \textbf{0.971} \\
RunningTitleZone    & \textbf{0.339} & 0.272 & 0.287 & 0.220 & 0.301 & 0.249 \\
StampZone           & 0.851 & 0.787 & 0.702 & 0.702 & 0.894 & \textbf{0.995} \\
TitlePageZone       & 0.500 & 0.995 & 0.995 & 0.995 & \textbf{1.000} &  \textbf{1.000} \\

\bottomrule
\end{tabular}}

\end{table}

On the more challenging and diverse CATMuS dataset, the ranking changes. \textbf{YOLOv11x-OBB (0.564 mAP)} emerges as the clear leader, significantly outperforming other models in the primary mAP metric and also achieving the highest mAP50:95. \textbf{Co-DETR (0.492)} secures the second position, demonstrating strong localization with the best AP75 (0.451). \textbf{Grounding DINO (0.452)} and \textbf{YOLO-World (0.425)} show a considerable drop compared to the e-NDP dataset and rank third and fourth, respectively. \textbf{YOLOv11x-AABB (0.360)} remains the least effective. Per-class results confirm YOLO-OBB's broader effectiveness, leading in 5 out of 9 common categories, particularly those benefiting from orientation modeling like DropCapitalZone, NumberingZone, and QuireMarksZone. Co-DETR excels at MainZone and TitlePageZone, while Grounding DINO leads in GraphicZone and StampZone.

\subsection{Results on HORAE Dataset (8 Classes)}
On the visually complex HORAE dataset (Table \ref{tab:overall_horae_final}), the trend observed in CATMuS is confirmed and amplified on the mAP50:95 metric. \textbf{YOLOv11x-OBB (0.568)} is again the top-performing model. \textbf{Co-DETR (0.523)} and \textbf{Grounding DINO (0.511 )} follow at a distance, while \textbf{YOLOv11x-AABB (0.458)} is less effective. Per-class results (Tables \ref{tab:per_class_map_horae_final} and \ref{tab:per_class_map50_horae_final}) show YOLO-OBB's strength in detecting distinct decorative elements like miniatures and initials.

\begin{table}[ht]
\centering
\caption{Overall Performance Comparison on HORAE Dataset (8 Classes)}
\label{tab:overall_horae_final}
\scalebox{1}{
\begin{tabular}{lccccc}
\toprule
Model /                  & mAP                     & mAP                    & Recall                 & Precision             & Speed \\
 Epochs                      & @.50:.95 ($\uparrow$)    & @.50 ($\uparrow$)      & (AR/R) ($\uparrow$)    & (P) ($\uparrow$)      & (\nicefrac{\text{im}}{\text{ms}}) ($\downarrow$) \\
\midrule
Grounding DINO (ep11)   & 0.511                & 0.703             & 0.615 (AR)      & N/A           & $\approx$115.0 \\
YOLOv11x-AABB (epXX)    & 0.458                & 0.665             & 0.600 (R)       & 0.646         & $\approx$6.9   \\
YOLOv11x-OBB (epXX)     & \textbf{0.568}       & \textbf{0.737}    & 0.597 (R)       & \textbf{0.736}    & $\approx$11.5  \\
YOLO-World x (epXX)$^*$ & 0.415                & 0.655             & 0.610 (R)       & 0.602         & $\approx$12.9  \\
Co-DETR (ep11)          & 0.523                & 0.735             & 0.640 (AR)      & N/A           & $\approx$150.0 \\
\midrule
YOLO-gen 11x-OBB (ep88)   & 0.505 & 0.650 & \textbf{0.666} (R) & 0.706 & $\approx$12.8  \\
\bottomrule
\end{tabular}}
\caption*{\footnotesize $^\dagger$Inference times (ms/im) on NVIDIA A6000.}
\end{table}

\begin{table}[htbp!]
\centering
\caption{Per-Class mAP@.50:.95 Comparison on HORAE Dataset}
\label{tab:per_class_map_horae_final}
\scalebox{0.98}{ 
\begin{tabular}{lcccccc}
\toprule
Models /                  & \small{Grounding}                     & \small{YOLO}                    & \small{YOLO}                 & \small{YOLO}             & \small{Co-DINO} & \small{YOLO-gen} \\
 Category                      & \small{DINO}    & \small{11x-AABB}      & \small{11x-OBB}    & Worldx (v2)  & \small{(co-DETR)} & \small{11x-OBB} \\
\midrule

Decorated Border    & 0.589 & 0.481 & \textbf{0.596} & 0.420 & 0.556 & 0.550 \\
Illustrated Border  & 0.201 & 0.106 & \textbf{0.244} & 0.150 & 0.127 & 0.168 \\
Miniature           & \textbf{0.932} & 0.797 & 0.920 & 0.750 & 0.839 & 0.840 \\
Decorated Initial   & 0.552 & 0.556 & \textbf{0.755} & 0.450 & 0.520 & 0.514 \\
Historiated Initial & \textbf{0.900} & 0.895 & \textbf{0.900} & 0.850 & 0.895 & 0.895 \\
Line Filler         & 0.580 & 0.522 & \textbf{0.666} & 0.480 & 0.566 & 0.593 \\
Simple Initial      & 0.331 & 0.286 & \textbf{0.429} & 0.310 & 0.354 & 0.268 \\
Border Text         & 0.001 & 0.017 & 0.206 & 0.100 & 0.151 & \textbf{0.214} \\

\bottomrule
\end{tabular}}
\end{table}

\begin{table}[htbp!]
\centering
\caption{Per-Class mAP@.50 Comparison on HORAE Dataset}
\scalebox{0.98}{ 
\label{tab:per_class_map50_horae_final}
\begin{tabular}{lcccccc}
\toprule
Models /                  & \small{Grounding}                     & \small{YOLO}                    & \small{YOLO}                 & \small{YOLO}             & \small{Co-DINO} & \small{YOLO-gen} \\
 Category                      & \small{DINO}    & \small{11x-AABB}      & \small{11x-OBB}    & Worldx (v2)  & \small{(co-DETR)} & \small{11x-OBB} \\
\midrule

Decorated Border    & 0.750 & 0.644 & \textbf{0.817} & 0.650 & 0.721 & 0.755 \\
Illustrated Border  & 0.359 & 0.240 & \textbf{0.428} & 0.280 & 0.307 & 0.251 \\
Miniature           & \textbf{1.000} & 0.953 & 0.965 & 0.910 & 0.958 & 0.878 \\
Decorated Initial   & 0.904 & 0.937 & \textbf{0.967} & 0.700 & 0.890 & 0.728 \\
Historiated Initial & \textbf{1.000} & 0.995 & \textbf{1.000} & 0.990 & 0.995 & 0.995 \\
Line Filler         & \textbf{0.980} & 0.965 & 0.949 & 0.720 & 0.964 & 0.888 \\
Simple Initial      & \textbf{0.629} & 0.543 & \textbf{0.629} & 0.550 & 0.659 & 0.410 \\
Border Text         & 0.003 & 0.042 & 0.257 & 0.150 & 0.263 & \textbf{0.291} \\

\bottomrule
\end{tabular}}
\end{table}

\section{Discussion}

\begin{figure}
\scalebox{0.47}{
\includegraphics{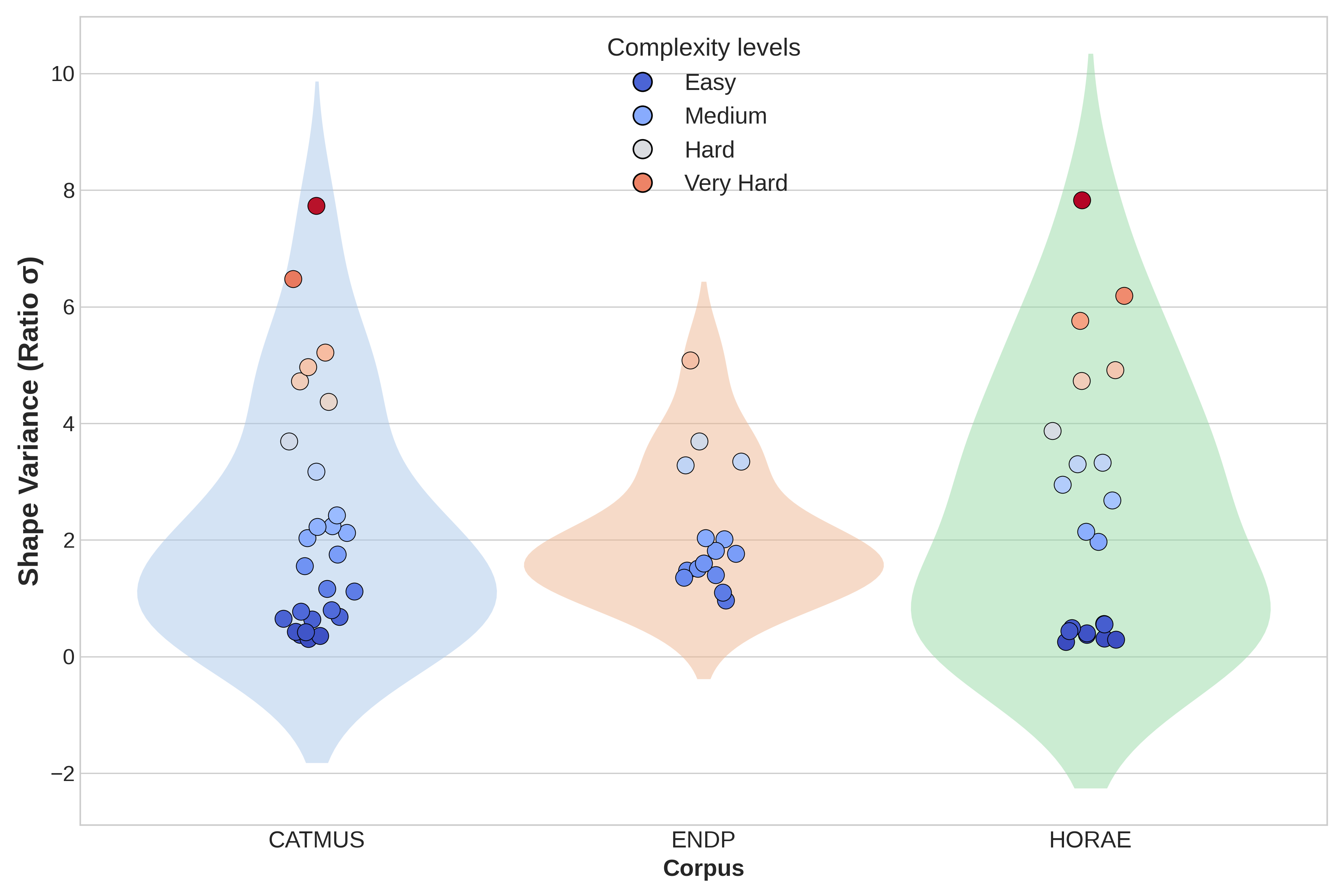}}
\centering
\caption[foo bar]{\small{Visual complexity profile based on Aspect Ratio (longer side / shorter side) on a logarithmic scale. Each class is represented by its mean, and its 25th/75th percentiles. The spread between points indicates the shape variability, confirming the higher heterogeneity of HORAE and CATMuS classes.}

}
\label{fig:complexity_Q1}
\end{figure}

The empirical results reveal a complex interplay between model architecture, dataset characteristics, and the underlying nature of the historical documents. Our findings indicate that model performance is highly contingent on the codicological complexity of the source material (See figure \ref{fig:complexity_Q1}), moving the discussion beyond simple metrics towards an evaluation of architectural suitability for specific historical contexts.

\textbf{Dataset-Contingent Performance and Architectural Strengths:}
The performance hierarchy of the tested models shifted significantly across the three corpora, demonstrating a clear sensitivity to data characteristics. 
\begin{itemize}
    \item On the \textbf{e-NDP} dataset, characterized by its functional, evolutive but highly structured \textit{mise-en-page}, Transformer-based models excelled. \textbf{Co-DETR} achieved the highest mAP (0.752), likely because its global attention mechanism is adept at learning the stable, long-range spatial relationships inherent in these administrative registers.
    \item Conversely, on the more diverse \textbf{CATMuS} and the visually complex \textbf{HORAE} datasets, the advantage shifted to \textbf{YOLOv11x-OBB}. As intra-class visual variance and layout complexity increased, the strong inductive bias of YOLO's convolutional layers, which excel at identifying local features, proved more resilient than the global approach of Transformers. This suggests that while Transformers may be superior for learning predictable structures, robust CNNs are better at generalizing across the high artistic and scribal variance found in more heterogeneous or ornate manuscript collections.
\end{itemize}

\textbf{The Codicological Imperative for Oriented Bounding Boxes (OBB):}
The experiments unequivocally demonstrate that using Oriented Bounding Boxes is not a minor refinement but a fundamental necessity for accurate DLA in this domain. Across datasets, the OBB variant of YOLO vastly outperformed its axis-aligned counterpart (e.g., a 0.564 mAP for YOLOv11x-OBB versus 0.360 for YOLOv11x-AABB on CATMuS). This technical superiority is a direct consequence of the physical and diplomatic reality of manuscripts. An AABB imposes a modern, Cartesian logic onto a non-Cartesian object, failing to account for:
\begin{enumerate}
    \item \textbf{Material Reality:} Parchment warps and cockles over time, and bindings introduce page curvature, meaning even intentionally straight lines are rarely perfectly axis-aligned.
    \item \textbf{Diplomatic Reality:} Scribes wrote at a natural angle, and decorative elements like vine-stem borders were designed with organic curves.
\end{enumerate}
An AABB, by definition, must encompass the outermost points of a skewed or curved element, thereby including large areas of background or overlapping with adjacent,unrelated elements. This introduces significant label noise during training and severe localization errors during inference. The OBB approach provides a far more faithful geometric representation, leading to superior localization and classification accuracy.

\textbf{Architectural Trade-offs and Research Implications:}
The choice of architecture has practical implications that transcend mAP scores.
\begin{itemize}
    \item \textbf{Transformers (Co-DETR, Grounding DINO):} These models offer a more holistic page understanding, making them suitable for tasks where global context is key. However, their data hungriness and slower inference make them a costly choice for massive-scale processing. Their performance degradation on complex datasets suggests a trade-off between contextual understanding and generalization robustness.
    \item \textbf{Vision-Language Models (Grounding DINO, YOLO-World):} While their fine-tuning performance was moderate, their primary value lies in the potential for few-shot and open-vocabulary detection. This enables a researcher to query for arbitrary, semantically-defined elements (e.g., 'manicules' or specific iconographic motifs) without model retraining, a paradigm-shifting capability for exploratory analysis. Their weaker performance in this closed-set benchmark suggests a trade-off : the generalized semantic knowledge struggles when forced to discriminate between visually similar, pre-defined classes (e.g., ‘Decorated Initial‘ vs. ‘Historiated Initial‘), where purely visual models excel.
    \item \textbf{CNN-OBB (YOLOv11x-OBB):} This architecture represents the most pragmatic tool for our specific task of high-accuracy, closed-set detection at scale. It is fast, robust, and its architectural priors (local convolutions + oriented detection head) are well-suited to the visual reality of manuscripts. Its main limitation is its semantic ignorance; it is a powerful pattern-matcher that requires a complete, new training cycle for any new set of layout classes.

    \item \textbf{The Foundational Model (YOLO-gen):} Our final unified model represents a new paradigm. By training on a harmonized, hierarchical dataset, it trades peak specialization for cross-domain versatility. While it may not achieve the absolute highest mAP on every specific subclass, its performance is highly competitive across all corpora. Crucially, its unique ability to detect abstract parent classes (Table \ref{tab:yolo_gen_parents}) with high Precision and Recall (>70\%) validates it as a powerful foundational tool, ideal as a starting point for new DLA projects.

\end{itemize}

\textbf{Limitations and Future Directions:}
This study has limitations that open avenues for future work. Our experiments used specific model variants; larger backbones or more exhaustive hyperparameter tuning, or different historical corpora might alter the results. 

A crucial issue, evident in Table \ref{tab:class_dist_compact}, is the severe \textbf{class imbalance} inherent in all corpora. Every dataset contains “majority” classes (e.g., ‘MainZone‘) and “long-tail” classes with very few instances (e.g., ‘TitlePageZone‘ ‘Historiated Initial‘). This biases models towards common classes and can lead to poor performance on rare categories that are often of great scholarly interest. Future work should explore methods to mitigate this, such as cost-sensitive learning, data augmentation strategies for rare classes, or few-shot learning approaches. 

Furthermore, our evaluation relied on bounding box metrics (Box mAP); extending the analysis to segmentation quality (Mask mAP) would provide deeper insights, especially for irregularly shaped decorative elements and for comparing against instance segmentation models like Mask R-CNN or multi-modal systems like Florence-2.

\section{Conclusion}

Our benchmark of state-of-the-art object detectors across three distinct manuscript corpora confirms that the optimal architecture is highly contingent on the codicological complexity of the source material. While Transformer-based models like \textbf{Co-DETR} excel on structured layouts, specialized CNNs with Oriented Bounding Boxes, particularly \textbf{YOLOv11x-OBB}, are superior for visually diverse, single-domain datasets. This work also validates that modeling on Oriented Bounding Boxes (OBB) is a fundamental requirement for accurate layout analysis in historical documents, consistently outperforming axis-aligned methods by faithfully representing the non-Cartesian and artistic nature of manuscripts.

Building upon these findings, one primary contribution of this research is the demonstration that data harmonization through a codicologically-informed, hierarchical ontology can produce a single, generalist model—\textbf{YOLO-gen}—with performance nearly on par with domain-specific specialists. This result challenges the prevailing notion that high performance requires narrow specialization. It proves that it is possible to overcome the challenge of disparate, non-interoperable datasets, validating a methodology that creates robust, foundational models for codicological analysis. 

This study highlights a critical trade-off for practitioners: while emerging Vision-Language Models offer exciting possibilities for zero-shot querying, their performance is currently surpassed by specialized architectures. For large-scale, the combination of a fast CNN backbone and an OBB detection head offers the most effective and reliable solution at present. Future work should focus on bridging this gap by integrating orientation-aware heads into VLMs and developing robust strategies for handling long-tail class distributions. This will provide a more holistic guide for scholars selecting computational tools to unlock the rich contents of our written heritage.

\section{Models and Datasets}

The datasets supporting this work are freely available on open repositories:
\vspace{5pt}

e-NDP:  \href{https://doi.org/10.5281/zenodo.7575693}{https://doi.org/10.5281/zenodo.7575693}
\vspace{7pt}

CATMuS: \href{https://huggingface.co/datasets/CATMuS/medieval-segmentation}{https://huggingface.co/datasets/CATMuS/medieval-segmentation}
\vspace{7pt}

HORAE: \href{https://github.com/oriflamms/HORAE}{https://github.com/oriflamms/HORAE}
\vspace{7pt}

YOLO-historical: \href{https://huggingface.co/datasets/magistermilitum/YOLO_historical}{https://huggingface.co/datasets/magistermilitum/YOLO\_historical}
\vspace{12pt}

YOLO models:
\vspace{7pt}

YOLO-gen : \href{https://huggingface.co/magistermilitum/YOLO_manuscripts}{https://huggingface.co/magistermilitum/YOLO\_manuscripts}



\bibliographystyle{plain}
\bibliography{references}

\end{document}